\documentclass{article}

\usepackage{tikz}
\usetikzlibrary{shapes,arrows,positioning}
\usetikzlibrary{calc}

\usepackage{array}
\usepackage{hyperref}
\usepackage{url}
\usepackage{multirow}
\usepackage{ulem}
\usepackage{graphicx}
\usepackage{caption}
\usepackage{csquotes}
\usepackage{float}
\usepackage{microtype}
\usepackage{subfigure}
\usepackage{booktabs} 
\usepackage{amsmath}
\usepackage{amssymb}
\usepackage{mathtools}
\usepackage{amsthm}
\usepackage[capitalize,noabbrev]{cleveref}

\usepackage[accepted]{icml2025}

\begin{document}

\twocolumn[
\icmltitle{JOOCI: a Novel Method for Learning Comprehensive Speech Representations}

\begin{icmlauthorlist}
\icmlauthor{Hemant Yadav}{1}
\icmlauthor{Rajiv Ratn Shah}{1}
\icmlauthor{Sunayana Sitaram}{2}\\

\end{icmlauthorlist}

\icmlaffiliation{1}{IIIT Delhi, India}
\icmlaffiliation{2}{Microsoft Research}

\icmlcorrespondingauthor{Hemant Yadav}{hemantya@iiitd.ac.in}

\icmlkeywords{Machine Learning, ICML, speech representation learning, SSL, SOTA, JOOCI, MPL, MMPL, HuBERT, WavLM}

\vskip 0.3in
]

\printAffiliationsAndNotice{}

\begin{abstract}

Information in speech can be categorized into two groups: Content (what is being said, such as linguistics) and Other (how it is expressed such as information about speaker and paralinguistic features). 
Current self-supervised learning (SSL) methods are shown to divide the model's representational-depth or layers in two, with earlier layers specializing in Other and later layers in Content related tasks.
This layer-wise division is inherently sub-optimal, as neither information type can use all layers to build hierarchical representations.
To address this, we propose JOOCI, a novel speech representation learning method that does not compromise on the representational-depth for either information type.
JOOCI outperforms WavLM by 26.5\%, and other models of similar size (100M parameters), when evaluated on two speaker recognition and two language tasks from the SUPERB benchmark, demonstrating its effectiveness in Jointly Optimizing Other and Content Information (JOOCI).
Code and models will be released publicly \href{https://github.com/}{upon publication}.


\end{abstract}

\section{Introduction}
\label{sec:intro}

\begin{figure}[t]
\centering
\resizebox{1.05\columnwidth}{!}{
\begin{tikzpicture}[
    node distance=0.5cm,
    task/.style={rectangle, rounded corners, align=center},
    group/.style={rectangle, rounded corners, text width=3.2cm, minimum height=1cm, align=center},
    arrow/.style={->, >=stealth, thick},
]

\node (speech) [task] {\includegraphics[width=0.5\columnwidth]{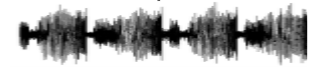}};

\node (content) [group, below right=0.5cm and -1cm of speech] {\textbf{Content} \\ (What is said)};
\node (task_c) [task, below=of content] {PR, ASR, ST};

\node (other) [group, below left=0.5cm and -1cm of speech] {\textbf{Other} \\ (How it is expressed)};
\node (task_o) [task, below=of other] {SID, SV, ER};

\draw [arrow] (speech) -- (content);
\draw [arrow] (speech) -- (other);
\draw [arrow] (content) -- (task_c);
\draw [arrow] (other) -- (task_o);

\end{tikzpicture}
}
\caption{Information in speech can be categorized into Other and Content. A model that claims to learn comprehensive speech representations, capturing both Other and Content information, must excel across downstream tasks, including ASR, PR, ST, SID, SV, and ER.}
\label{fig:speech_tasks}
\end{figure}

Self-supervised learning (SSL) methods have shown strong performance across different modalities, including text \citep{brown2020languagegpt3}, vision \citep{alexey2020imagevit}, and audio \citep{baevski2020wav2vec2.0, mohamed2022self, defossez2022highencodec}. 
This work focuses on the audio modality, specifically on learning high-level representations from speech data.
Information in speech can be categorized into Content (what is being said, such as linguistic content) and Other\footnote{In this work, we request the readers to keep in mind that we keep referring \textbf{Other} features/information using this definition.}  (how it is expressed such as information about speaker and paralinguistic features). 
As illustrated in Figure \ref{fig:speech_tasks}, robust performance across diverse downstream tasks such as speaker verification (SV) and phoneme recognition (PR) requires models to encode both information types effectively.
Thus the ultimate goal of a speech SSL method is to learn and encode a wide range of comprehensive and meaningful representations from speech data.
Therefore, a single model's ability to jointly optimize Other and Content is crucial for robust performance across diverse downstream tasks. 
However, existing SSL methods struggle to optimize these jointly, improving one often degrades the other \citep{baevski2022efficientdata2vec2.0, shi2023explorationMRHUBERT}.

\begin{figure*}[t]
\centering
\includegraphics[trim=2.75cm 3.4cm 2.75cm 3.4cm, clip,width=2\columnwidth]{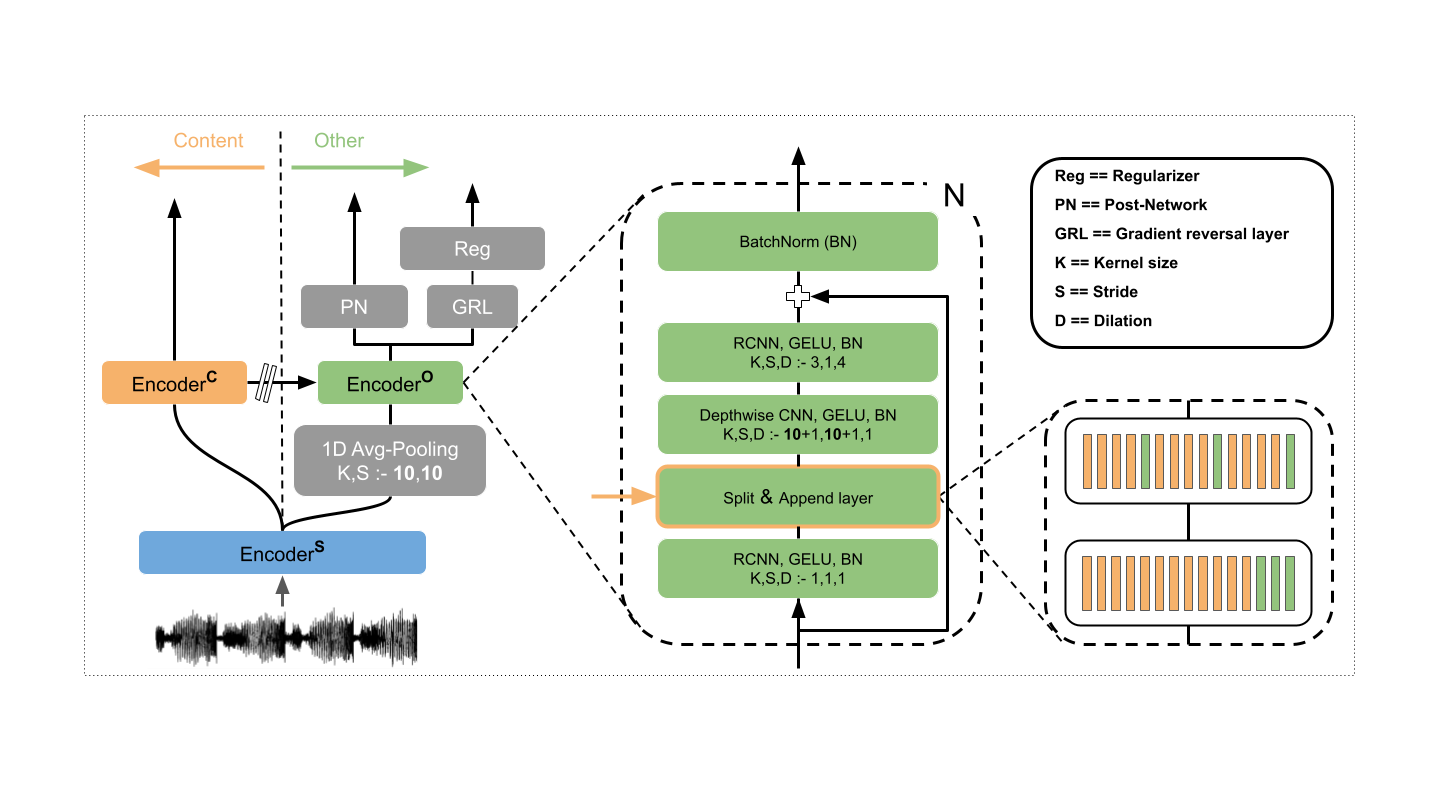}
\caption{JOOCI Method. 
As shown, raw audio is processed through the shared Encoder$^S$ ($E^S$), and the output is passed to the Encoder$^O$ ($E^O$) and Encoder$^C$ ($E^C$) encoders. The split-and-append mechanism enables the Other encoder to extract useful information from the Content encoder during the forward pass, while preventing gradient flow during the backward pass. The overall design ensures that both encoders can operate independently while still benefiting from $E^S$. During inference, $E^O$ is used for tasks requiring Other information and $E^C$ for Content related tasks.}
\label{figure:method}
\end{figure*}

The SUPERB benchmark \citep{yang2021superb} highlights this challenge, evaluating speech models across diverse downstream tasks such as automatic speech recognition (ASR), phoneme recognition (PR), speech translation (ST), speaker identification (SID), speaker verification (SV), and emotion recognition (ER). 
State-of-the-art (SOTA) approaches like WavLM \citep{chen2022wavlm} and HuBERT \citep{hsu2021hubert} on the SUPERB benchmark use masked prediction loss (MPL) and a transformer encoder architecture. While these methods have achieved significant success, there remains room for further improvement as explained below. 
 
Studies have shown that MPL, in HuBERT, tends to prioritize Content information at the expense of Other information, such as speaker identity \citep{yadav2023analysing} \footnote{In generative speech research, SSL methods are used to extract semantic (Content) tokens and neural codecs handle acoustic (Other) tokens. For instance, AudioLM \citep{borsos2023audiolm} uses w2v-BERT \citep{chung2021w2vbert} for Content tokens and SoundStream \citep{zeghidour2021soundstream} for Other tokens.}. For example, \citep{feng2022silenceissweeter} showed that, models pre-trained with MPL encode speaker (Other) information in the silent parts of audio, corroborating the earlier work that MPL, alone, may fall short in maximizing Other information. 

To overcome this, UniSpeech-SAT \cite{chen2022unispeechsat} uses a dedicated loss function combined with data augmentation for learning speaker (Other) information, but this led to reduced performance on Content related tasks.
To mitigate this, WavLM incorporated data augmentation alongside MPL and an architectural change, which has been shown to improve performance on speaker (Other) tasks while maintaining performance on Content tasks. 
However, both WavLM and earlier methods are shown to divide the model's total representational-depth or layers in two, with earlier layers specializing in Other and later layers in Content related tasks such as SV and PR respectively \citep{chen2022wavlm}.
We argue that this layer-wise division is inherently sub-optimal for learning comprehensive speech representations, as it prevents either information type from leveraging the full representational-depth of the model. Maximizing one leads to minimizing another.
For example, MS-HuBERT \citep{yadav2024ms-hubert} prioritizes performance on tasks such as PR by dedicating a disproportionate number of layers for modeling Content information, which worsens performance on speaker related tasks, further illustrating this inefficiency.

JOOCI is designed to address the limitations of existing SSL methods by jointly optimizing Other (how it is expressed) and Content (what is said) information without compromising representational-depth for either. Unlike prior methods \citep{chen2022wavlm} that divide the model’s layers into specialized groups, JOOCI employs separate encoders for Other and Content information while maintaining a shared encoder for efficient low-level feature extraction. 
Representational-depth provides significant advantages in learning, as shallow networks require exponentially more parameters to achieve comparable representational power \citep{Goodfellow-et-al-2016, resnet}. 
This architecture ensures that both types of information can leverage the full representational depth of the model.

Our contributions can be summarized as follows:
\begin{enumerate}

\item We present a novel speech SSL method, JOOCI, which uses separate learnable parameters and distinct loss functions, unlike WavLM and UniSpeech-SAT.
\item We evaluate JOOCI on two Content and two Other (speaker) tasks from the SUPERB benchmark, achieving a 26.5\% improvement over WavLM.
\item Code and models will be released upon publication.


\end{enumerate}

The paper is structured as follows: Section \ref{sec:relatedworks} reviews related work, highlighting current challenges in jointly modeling Other and Content information. Section \ref{sec:background} and \ref{sec:jooci} adds the background and introduces JOOCI’s architecture and training. Section \ref{sec:results} covers experimental setup and presents main results demonstrating JOOCI’s superiority. Section \ref{sec:ablation} offers an discussion of JOOCI with Adapters, followed by conclusions and future directions.

\section{Related work}
\label{sec:relatedworks}
Self-supervised learning (SSL) models have driven remarkable progress across multiple domains, including text \citep{brown2020languagegpt3, touvron2023llama1, dubey2024llama3}, vision \citep{alexey2020imagevit, rombach2022highstablediffsion}, and speech \citep{schneider2019wav2vec, chung2021w2vbert, baevski2020wav2vec2.0, hsu2021hubert, baevski2022efficientdata2vec2.0}. In the speech domain, SSL methods aim to learn comprehensive representations that encode both Content (what is said, e.g., linguistics) and Other information (how it is expressed, e.g., speaker identity). However, prior SSL models have shown a tendency to divide the model’s total representational-depth or layers into two, with earlier layers specializing in Other information and later layers focusing on Content-related tasks such as speaker verification (SV) and phoneme recognition (PR), respectively \citep{yadav2023analysing, feng2022silenceissweeter}.

Masked prediction loss (MPL), widely used in SSL models \citep{hsu2021hubert, baevski2022data2vec}, is particularly effective for Content tasks like PR and ASR. The choice of loss function plays a crucial role in shaping the learned representations, whether for Content or Other information \citep{audiodivide, wang2018additiveamsoftmax, mohamed2022self}.
UniSpeech-SAT \citep{chen2022unispeechsat} introduced contrastive speaker-aware loss and data augmentation alongside MPL, improving speaker and emotion tasks at the cost of reduced performance on linguistic tasks. Similarly, applying MPL at multiple resolutions enhances linguistic representations but degrades speaker-related performance \citep{shi2023MRHuBERT, yadav2023analysing}.
Both the observations can be explained because the total representational-depth of the encoder is disproportionally used for either modeling Other or Content information. 
WavLM uses MPL with only data augmentation and outperformed HuBERT across various downstream tasks requiring effective modeling of Other and Content information. For speaker tasks UniSpeech-SAT still performed a little better showing the importance of using separate loss function. These findings highlight the need for separate learnable parameters and loss functions to effectively model both Content and Other information.

Inspired by these findings, JOOCI is designed to address the inherent limitations of prior SSL models by jointly optimizing Content and Other information without compromising representational-depth. Unlike previous methods that statically allocate model layers to different tasks, JOOCI employs separate encoders with a shared feature extraction module, allowing both types of information to fully utilize the network’s total representational capacity and separate loss functions. Our experiments on the SUPERB benchmark validate this approach, showing significant improvements over WavLM in both Content and speaker (Other) related tasks.

\textbf{What JOOCI is not?} Unlike models that aim to maximize Content information while suppressing Other \citep{yadav2024ms-hubert, shi2023MRHuBERT, qian2022contentvec, chang2023selfspin, chang2023selfspin2}, JOOCI does not prioritize one type of information over another. Also, JOOCI is not a disentanglement method similar to \citep{zhao2023ccsrddisentangel, chan2022contentdisentable2}. Instead, JOOCI is designed to learn comprehensive speech representations that capture both linguistic and paralinguistic attributes without trade offs in representational-depth. 


\section{Background}
\label{sec:background}
\subsection{MS-HuBERT}
MS-HuBERT \citep{yadav2024ms-hubert} uses Multicluster MPL to improve the performance of Content related tasks such as ASR and PR. It follows an iterative pre-training SSL approach, similar to HuBERT, and consists of two encoders: a convolutional neural network (CNN) encoder (1st), followed by a transformer encoder (2nd). The CNN encoder serves a dual purpose, including the down-sampling of input data. During pre-training, raw audio is passed through the CNN encoder, and approximately $50\%$ of the output tokens are masked, using the masking token $[M]$before being fed into the transformer encoder. The network is then trained to optimize to output a discrete target sequence by minimizing the Multicluster MPL. 
JOOCI uses MS-HuBERT to model Content information. For further details related to MS-HuBERT, please refer to the original paper.

\subsection{RDINO}
The Regularized Distillation Framework (RDINO) \citep{chen2023pushing} leverages the DINO (Distillation with NO labels) framework to learn robust speaker representations using SSL. By combining advance data augmentation techniques such as WavAugment and SpecAugment with novel regularization strategies, RDINO performed better than prior SSL methods and significantly narrowed the gap with fully supervised systems on the SV task. 
JOOCI uses RDINO as a teacher to train the Other encoder. For further details realted to RDINO, please refer to the original paper.

\section{Method}
\label{sec:jooci}

Unlike prior SSL methods, JOOCI uses two separate encoders while maintaining a shared encoder for efficient low-level feature extraction. This architecture ensures both information types fully utilize the available representational-depth to maximize performance.

Figure \ref{figure:method} illustrates the JOOCI architecture and Table \ref{table:param} shows the parameter count of each component. We detail the key components of JOOCI and their roles in the overall framework in the following section. 

\subsection{Components}
\textbf{Shared Encoder (E$^S$) :} The shared encoder downsamples raw audio into 20 ms segments using a 7-layer CNN (320$\times$ downsampling), identical to MS-HuBERT's CNN encoder (1st) \citep{yadav2024ms-hubert}. These embedding are input for both the Content and Other encoders. 

\textbf{Content Encoder (E$^C$) :} The Content encoder is responsible for maximizing linguistic information, such as phonemes and words. It adopts a transformer architecture with self-attention layers, identical to MS-HuBERT's transformer encoder (2nd). 
During pre-training, approximately 50\% of the input embeddings from E$^S$ are masked and replaced with the masked embedding $[M]$.

\textbf{Other Encoder (E$^O$) :} The Other encoder is responsible for maximizing non-linguistic information, such as speaker.
The embeddings from E$^S$ are further downsampled using a 1D average pooling layer (kernel size and stride equals to $10$), resulting in one embeddings for each 200 ms segment.
Each Other encoder block consists of the following:
\begin{itemize}
    \item 1D Res2Net block \citep{gao2019res2net} with a kernel size, stride, and dilation all equal to 1.
    \item A split-and-append layer which allows embeddings from the Content encoder to be combined with the Other encoder, enabling cross-information sharing.
    \item A depthwise 1D CNN layer \footnote{The number of groups equal to the Other embedding dimension.} with a kernel size and stride equal to 11 to map the length back 
    \item 1D Res2Net block with a kernel size, stride and dilation equal to 3, 1, and 4, respectively.
    \item A residual connection, followed by a 1D batch normalization (BN) layer.
\end{itemize} 

To enable the Other encoder to leverage useful information from the Content encoder, JOOCI uses a simple split-and-append layer. During the forward pass, embeddings from the Content encoder are split into groups of 10, and embeddings from the Other encoder are appended sequentially to these groups one by one at the end of each group sequentially, increasing the total total length of the input. To map it back to its original length, a depthwise CNN layer is applied with a kernel size and stride equal to $10+1$. Lastly, during the backward pass, gradients do not flow from the Other encoder to the Content encoder, maintaining their distinctiveness.

\textbf{Post Network :} The Post Network is used for training the Other encoder using the student-teacher framework, aligning its output with that of RDINO model (teacher) through the following steps:
\begin{itemize}
    \item An attentive statistical pooling (ASP) layer, similar to \citep{okabe2018attentiveasp}, which aggregates variable-length inputs. 
    \item A 1D batch normalization (BN) layer. 
    \item A fully connected (FC) layer with an output dimension of 512. This layer is used only during the pre-training step.
\end{itemize}

\textbf{Regularizer :} The role of regularizer network is to prevent the Other encoder from over-specializing in speaker-related information. 
It is a one layer transformer decoder with 8 heads and a 768-dimensional hidden size with a gradient reversal layer (GRL) applied before back-propagating gradients to the Other encoder. 
We found that it is crucial improving performance on tasks like emotion recognition (ER) and is only used during pre-training.


\renewcommand{\arraystretch}{1.1}
\begin{table}[t]
    \caption{Parameter count for different components of the JOOCI method in millions.}
    \centering
    \scalebox{1.0}{
    \begin{tabular}{c|c|c}
    \hline
    \textbf{-} & \textbf{Training} & \textbf{Inference} \\
    \hline
    HuBERT & 94.68M & 94.68M \\
    \hline
    WavLM & 94.70M & 94.70M \\
    \hline
    \multicolumn{3}{c}{ JOOCI (ours)} \\
    \hline
    Shared \& Content encoder  & 96.18M & 94.68M \\
    \hline
    Other encoder & 3.12M & 3.12M \\
    \hline
    Post network & 0.40M & - \\
    \hline
    Regularizer & 9.3M  & - \\
    \hline
    JOOCI (Ours) & 109M & 97.80M \\
    \hline
    \end{tabular}
    }
\label{table:param}
\end{table}

\subsection{Training Objective}

JOOCI is trained using three loss functions.

\textbf{Content Loss (CL):} This loss is computed using Multicluster Masked Prediction Loss (MMPL), which is shown to maximize Content information learnt. The MMPL, used by \citep{yadav2024ms-hubert}, calculates the MPL across multiple layers of the Content encoder using six sets of different pseudo labels. These layers are selected at regular intervals between the last and an intermediate layer. MMPL computes the sum of MPL over a set $d$ as shown in Equation \ref{eq:MMPL}.

\begin{equation}
\label{eq:MMPL}
    \mathcal{L}_{CL} = \sum_{d}^{} (MPL) 
\end{equation}
where $d$ is a dictionary indicating which pseudo label set corresponds to which Content encoder layer. 

The MPL is computed, similar to HuBERT, only at the masked indices, as shown in Equation \ref{eq:MPL}.

\begin{equation}
\label{eq:MPL}
\mathcal{L}_{MPL} = \frac{1}{\mathcal{M}} \sum_{i=1}^{\mathcal{M}} \frac{ exp \ (\text{sim}(Ah_i, \mathbf{e}_c) / \tau )} {\sum_{c'=0}^{C-1} exp \ (\text{sim}(Ah_i, \mathbf{e}_{c'}) / \tau )}
\end{equation}

Here, $\mathcal{M}$ is the masked indices, $A$ is the projection matrix, $h_i$ is the Content encoder embedding, $e_c$ and $e_{c'}$ represent the correct and incorrect embeddings for the pseudo labels, $sim(·, ·)$ computes cosine similarity between two vectors, and $\tau$ is a scaling factor for the logits. For further details, please refer to \citep{yadav2024ms-hubert, hsu2021hubert}.

\textbf{Other Loss (OL) :} The Other encoder is trained using a student-teacher framework, where RDINO is the teacher. RDINO has 22.74 million parameters, significantly more than the 3.52 million parameters in the Other encoder. The goal is to maximize the cosine similarity between the embeddings produced by the Other encoder (student) and RDINO (teacher), ensuring robust speaker representations. For each utterance, we use RDINO (teacher) to extract 512-dimensional embeddings for training the Other encoder (student) using the student-teacher framework. The full objective is shown in Equation \ref{eq:CS}.

\begin{equation}
\label{eq:CS}
\mathcal{L}_{\text{OL}} = 1 - \text{sim}(\footnotesize {Student, Teacher})
\end{equation}

Similar to the BYOL approach \citep{grill2020bootstrapbyol} we only use the positive pairs for loss calculation. 

\textbf{Regularizer Loss (RL) :} Similar to \citep{ao2022prespeech2c}, the decoder is trained to predict pseudo labels, using cross-entropy loss as defined in Equation \ref{eq:CE}. The pseudo labels, with a vocabulary size of 1005, are cluster centers generated via K-means and used in MMPL. Since these clusters also capture non-linguistic information, the combination with a GRL layer enables their use as a regularizer.
\begin{equation}
\label{eq:CE}
\mathcal{L}_{\text{RL}} = - \sum_{t=1}^{T} \sum_{i=1}^{V} y_{t,i} \log \hat{y}_{t,i}
\end{equation}

Where $T$ is the total number of time steps , $V$ is the size of the vocabulary, $y_{t,i}$ and $\hat{y}_{t,i}$ are the true one-hot encoded pseudo labels and predicted probability for the $t^\text{th}$ time step and $i^\text{th}$ word in the vocabulary respectively.

\textbf{Total Loss :} The overall training objective is the sum of all individual losses, as shown in Equation \ref{eq:total}. 

\begin{equation}
\label{eq:total}
    \mathcal{L}_{Total} = \mathcal{L}_{CL} + \mathcal{L}_{OL} + \frac{\mathcal{L}_{RL}}{10}
\end{equation}

The $\mathcal{L}_{CL}$ term does not influence the parameters of the Other encoder, and the $\mathcal{L}_{OL}$ and $\mathcal{L}_{RL}$ terms do not affect the parameters of the Content encoder. Since JOOCI uses separate learnable parameters for Other and Content encoders, all the losses can be simply summed directly, except the regularizer and does not require careful hyperparameter tuning.

\section{Experiments}
\label{sec:results}
\renewcommand{\arraystretch}{1.1}
\begin{table*}[ht]
    \centering
    \caption{Evaluation of speech representation. To provide readers with a broader perspective of how performance scales with data size, we also include results for WavLM+, which has been trained on  $\sim100$ times more data. \textbf{However, WavLM+ and WavLM Large should not be directly compared to JOOCI, as the pre-training data (corpus) disparity makes such comparisons inappropriate.}}
    \scalebox{1.0}{
    \begin{tabular} {l|c|c|c|c|c|c|c}
    \hline
       \multicolumn{1}{c|}{\multirow{3}{*}{\textbf{Method}}} & \multicolumn{1}{c|}{\multirow{3}{*}{\textbf{\#Params}}}  &  \multicolumn{1}{c|}{\multirow{3}{*}{\textbf{Corpus}}} &  \multicolumn{2}{c|}{\multirow{1}{*}{\textbf{Speaker}}} & \multicolumn{2}{c|}{\multirow{1}{*}{\textbf{Content}}} & \textbf{Overall} \\

       & & & \textbf{SID} & \textbf{ASV} & \textbf{PR} & \textbf{ASR} & \textbf{Score} \\
       \cline{4-8}
        & & & Acc $\uparrow$ & EER $\downarrow$ & PER $\downarrow$ & WER $\downarrow$ & $\downarrow$ \\
        \hline
        FBANK & 0.00M & - & 8.5e-4 & 82.01 & 9.56 & 23.18 & 53.69 \\
        \hline
        modified CPC \citep{modifiedcpc} & 1.84M & LL 60k hr & 39.63 & 12.86 & 42.54 & 20.18  & 33.99\\
        \hline
        \hline
        wav2vec 2.0 \citep{baevski2020wav2vec2.0} & 95.04M & LS 960 hr & 75.18 & 6.02 & 5.74 & 6.43  & 10.75 \\
        \hline
        HuBERT \citep{hsu2021hubert} & 94.68M & LS 960 hr & 81.42 & 5.11 & 5.41 & 6.42  & 8.88 \\
        \hline
        WavLM \citep{chen2022wavlm} & 94.70M & LS 960 hr & 84.51 & 4.69 & 4.84 & 6.21  & 7.81 \\ 
        \hline
        MS-HuBERT \citep{yadav2024ms-hubert} & 96.01M & LS 960 hr & - & - & 4.17 & 5.32 & - \\
        \hline
        \textbf{JOOCI (Ours)} & 109.00M & LS 960 hr & \textbf{90.79} & \textbf{4.15} & \textbf{4.25} & \textbf{5.35}  & \textbf{5.74} \\
        \hline
        \hline

        WavLM + \citep{chen2022wavlm} & 94.70M & Mix 94k hr & 89.42 & 4.07 & 3.92 & 5.59 &  6.04 \\ 
        \hline
        WavLM Large \citep{chen2022wavlm} & 316.62M & Mix 94k hr & 95.49 & 3.77 & 3.06 & 3.44 &  3.70 \\ 
        \hline

        \end{tabular}}
\label{table:superb}
\end{table*}

\subsection{Experimental setup}
\label{sec:experimentalsetup}
\textbf{Pre-training Dataset :} For pre-training JOOCI, we use the LibriSpeech 960-hour dataset \citep{librispeechdataset}, excluding transcriptions, in line with prior studies in speech representation learning \citep{schneider2019wav2vec, baevski2020wav2vec2.0, chung2021w2vbert}.

LibriSpeech is a read speech dataset recorded in a clean environment, free from real-world noises, and contains under 1000 hours of data. We found that applying weakly data augmentation helps in speaker and emotion (Other) tasks without the loss of performance on Content related tasks.
We augment 12.5\% of the audio samples using the MUSAN corpus \citep{snyder2015musan} with a high signal-to-noise ratio (SNR) in the ranges between [5, 15] for noise, [13, 20] for speech, and [5, 15] for music. For comparison, WavLM, applies strong data augmentation, augments 50\% of the audio with a SNR value in between [-5, 5]. Lastly, we add Room Impulse Response (RIR) for reverberation to the selected audio samples, similar to RDINO.

\textbf{Pre-training :} Similar to MS-HuBERT base, we use 12 layers for the Content encoder and same number for the Other encoder. The Other encoder is trained using a student-teacher framework with RDINO \cite{chen2023pushing}, while the Content encoder is trained with six sets of Multicluster MPL, identical to MS-HuBERT \citep{yadav2024ms-hubert}. 

Pre-training involves two steps. Firstly, we freeze the Content encoder and train JOOCI for 50,000 iterations on 4 GPUs, with each GPU processing up to 375 seconds of audio, resulting in a total of 1500 seconds of audio per iteration. The learning rate is set to 5e-4, using the first 5,000 steps for warm-up. All other settings are consistent with those used in MS-HuBERT \cite{yadav2024ms-hubert}. 

Next, we unfreeze the Content encoder and continue training for an additional 100,000 iterations, using a learning rate of 5e-5 across 6 GPUs, with each GPU handling up to 200 seconds of audio, yielding a total of 1200 seconds of audio per iteration. The first 1,000 steps are used for warm-up updates. All other settings are consistent with those used in MS-HuBERT \cite{yadav2024ms-hubert}.

\textbf{Shared Encoder and Content Encoder Initialization : } The shared and Content encoder are initialized with the pre-trained MS-HuBERT model weights \citep{yadav2024ms-hubert} to save compute. This choice is motivated by MS-HuBERT's superior performance on Content-related tasks such as ASR and PR. 

\textbf{SUPERB :} The SUPERB benchmark is designed to compare models on their ability to learn comprehensive speech representations across a variety of downstream tasks including speaker recognition, linguistic analysis, semantics, para-linguistics, and generation. In the SUPERB benchmark the encoder is frozen and a weighted-sum of all the layers is learned to be used as representations for different downstream tasks. 
We evaluate JOOCI on two Content related tasks, PR and ASR using $E^C$, and two Other tasks, SID and ASV using $E^O$, from the SUPERB benchmark to assess its performance.

These tasks exhibit a strong inverse correlation across different SSL methods, where improving one negatively impacts the other. This challenge highlights the difficulty in jointly optimizing Content and Other information. By addressing this trade-off, a method can demonstrate its ability to learn comprehensive speech representations. Thus, optimizing both tasks would validate JOOCI’s effectiveness in balancing these competing objectives.

\textbf{Supervised Fine-tuning :} No supervised fine-tuning is applied. Following pre-training, the regularizer and fully connected layer in the Post Network are discarded. All JOOCI parameters are frozen after the pre-training for evaluations.

\subsection{Main Results}
\label{sec:mainresults}
Table \ref{table:superb} summarizes the performance of JOOCI. The overall score is the mean of all the four tasks after converting the accuracy to error i.e., $100-acc$. The results clearly show that JOOCI outperforms WavLM on all tasks. In particular, JOOCI is able to jointly maximize the performance on speaker (Other) and linguistic (Content) tasks. 

One could argue that JOOCI's use of RDINO as a teacher, which is trained using 2500 hours of unsupervised data, is the reason for its exceptional performance on speaker realted tasks. However, when compared to WavLM+, trained on 94000 hours of data (versus JOOCI's 960 + 2500 hours), the comparable performance on SID (with slight improvement) and ASV (with slight degradation) suggests that the gains are not solely due to data volume but reflects JOOCI's ability to use the full representational depth for joint optimization of Other and Content information. Only for reference, WavLM Large with its 24 layers, further corroborates this theory that not just more data but higher representational depth is also crucial for learning comprehensive speech representations.

\renewcommand{\arraystretch}{1.2}
\begin{table*}[t]
    \caption{Performance of JOOCI's Other encoder on the SID task vs different combination layers used. For comparison the performance of HuBERT when using only the last layer is  $69.78$ compared to JOOCI's $90.24$ as shown  by \citep{yang2024large}}
    \centering
    \scalebox{1.05}{
    \begin{tabular}{c|c|c}
    \hline
    \textbf{Layers} & \textbf{Acc} $\uparrow$ & \textbf{Importance of other encoder layers} \\
    
    \hline
    \hline
    \multicolumn{3}{c}{Weighted-sum} \\ 
    \hline
    13 & 88.84 & \multirow{3}{*}{\includegraphics[trim=1.4cm 0.5cm 2.2cm 2.4cm, clip,width=0.6\columnwidth]{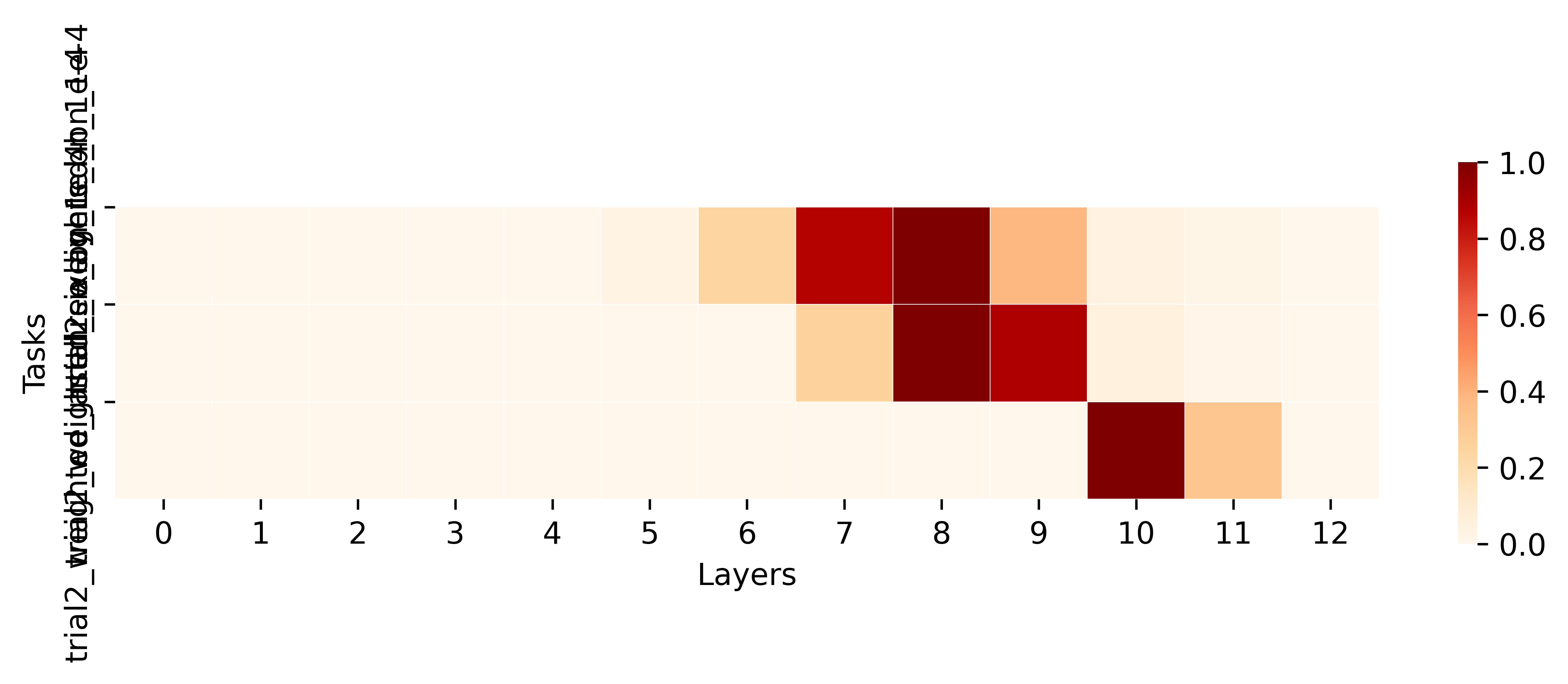}}  \\ 
    6 & 89.10&  \\
    
    3 & 88.87 &  \\ 
      &  & \\
    \hline
    \hline
    \multicolumn{3}{c}{Single Layer} \\ 
    \hline
    
    LastLayer & 90.24  & \multicolumn{1}{c}{\multirow{5}{*}{\textbf{Not Applicable}}} \\
    + ASP & 89.49  \\
    + BN & 90.79  \\
    + FC & 88.80  \\
    \hline
    \hline
    
        \end{tabular}}
    
\label{table:ablation_sidlayers}
\end{table*}

\begin{figure}[ht]
\centering
\includegraphics[,width=1\columnwidth]{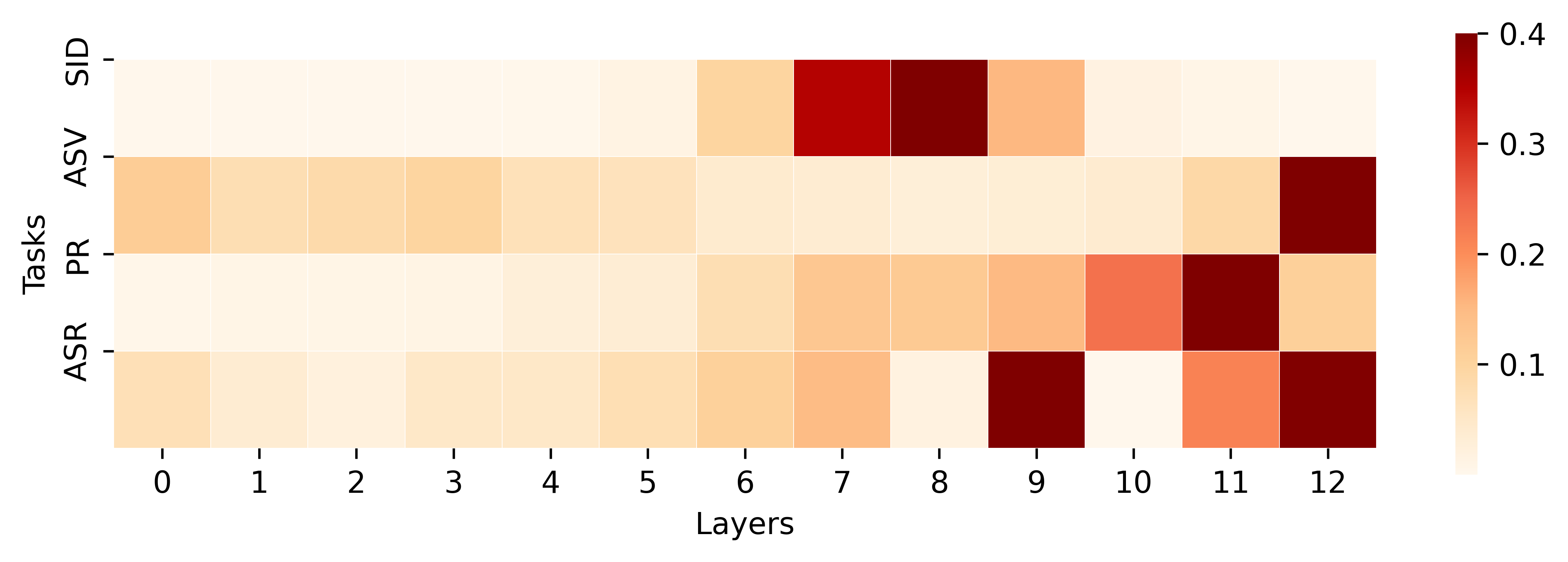}
\caption{Weight analysis on the SUPERB benchmark. Layer 0 corresponds to the input of the first Transformer layer. The y-axis represents different tasks, while the x-axis represents different layers. The higher the layer weight, the greater its contribution to the weighted sum.
For comparison with HuBERT and WavLM please see appendix Section \ref{section:hubwavsuperb}.}
\label{figure:heatmap}
\end{figure}

\subsection{Evaluating the performance on the SID task vs Information Encoded at Different Layers in the Other Encoder}

Table \ref{table:ablation_sidlayers} shows the accuracy for the SID task when using different combination of layers.
When using only the last layer gives the best result. Next, applying an ASP layer followed by Batch Normalization (BN) layer to the output of the last layer boosts the performance even further. On the other hand, using the last fully connected (FC) layer resulted in a performance drop of approximately ~2\%. This performance decline when using the last layer, closest to the loss function, has been observed in vision model trained using SSL, where it is often recommended to drop the last layer after pre-training \citep{chen2020simplesimclr}. These results prove that JOOCI's Other encoder is using the full available representational-depth for building hierarchical representation, unlike WavLM or HuBERT. Furthermore, as shown in Figure \ref{figure:heatmap}, JOOCI is able to jointly optimize Other and Content information without compromising on representational-depth for either.

\begin{figure}[ht]
\centering
\includegraphics[width=1\linewidth]{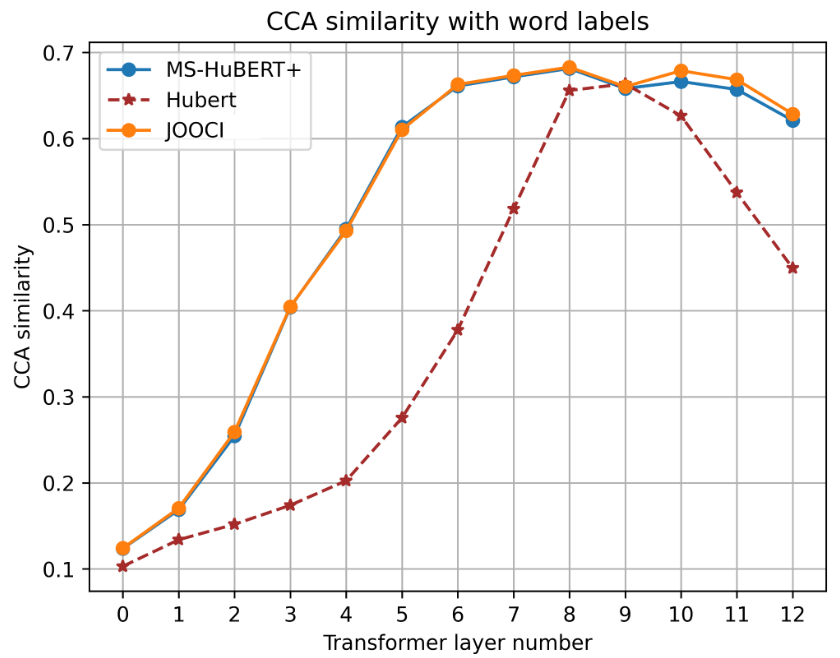}
\caption{Studying the effect of data augmentation on the content encoder using CCA word label similarity. Higher the CCA similarity for more number of layers better the method is.}
\label{figure:cca-similarity}
\end{figure}

Surprisingly, when using a weighted-sum of all the layers of Other encoder, the layers towards end (6-9) are assigned higher weight and not the last layer. To investigate this further, we use only the last six layers, without any performance drop, yet the same trend persisted. Next, we tried using only the last three layers, without any performance decline, which again showed the same pattern. This could be because of the magnitude of representations from different layers and we leave it for future work.

\subsection{Ablation: Understanding the Effect of JOOCI's different Component and Training Methodology}

Table \ref{table:ablation_grl} presents an ablation study analyzing the impact of JOOCI’s key components and training strategies on downstream tasks.
SID and ER task benefit a lot from data augmentation compared to the loss of performance on the ASR and PR. Similarly ER task benefits a lot by regularization compared to degradation on the SID task. 
Surprisingly, for the ER task content encoder embeddings with data augmentation performs better than Other encoder embeddings and worse without data augmentation. We leave this for future work.

\renewcommand{\arraystretch}{1.2}
\begin{table}[t]
    \centering
    \scalebox{1.0}{
    \begin{tabular}{c|c|c|c|c}
    \hline
       \multicolumn{1}{c|}{\multirow{2}{*}{\textbf{Encoder}}} & \textbf{SID} & \textbf{PR} & \textbf{ASR} & \textbf{ER} \\
       
        & Acc $\uparrow$ & PER $\downarrow$ & WER $\downarrow$ &Acc $\uparrow$ \\
        \hline
        JOOCI - $E^C$ & - & 4.25 & 5.35 & 65.27 \\
        - w/o data aug & - & 4.17 & 5.32 & 62.05 \\
        JOOCI - $E^O$ & 90.79 & - & - & 64.38 \\
        - w/o data aug & 88.50 & - & - & 62.72 \\
        - w/o reg & 88.94 & - & - & 61.86 \\
        
        \hline
        
        \end{tabular}}
    \caption{Effect of JOOCI's different Component and training methodology on downstream tasks performance.}
\label{table:ablation_grl}
\end{table}

\renewcommand{\arraystretch}{1.2}
\begin{table}[h]
    \centering
    \caption{Performance of JOOCI when only using layers with high CCA similarity score.}
    \scalebox{1}{
    \begin{tabular}{c|c|c}
    \hline
    \textbf{Method} & \textbf{PR} & \textbf{ASR}\\
    \hline
    HuBERT & 5.41 & 6.42 \\
    \hline
    WavLM & 4.84 & 6.21 \\
    \hline
    MS-HuBERT & 4.17 & 5.32  \\
    \hline
    JOOCI & 4.25 & 5.35  \\
    \hline
    \hline
    MS-HuBERT (6 - 11) & 4.05 & 5.25  \\
    \hline
    JOOCI (6 - 11) & 4.19 & 5.20  \\
    \hline

    \end{tabular}}
    
    \label{table:redovc}
\end{table}

\section{Discussion}
\label{sec:ablation}
\subsection{Effect of Data Augmentation on the Content Encoder embeddings} 
We investigate the effect of data augmentation on the nature of learned representations in the Content encoder. Following the approach of \citep{pasad2021layercpc, pasad2023comparative}, we plot Canonical Correlation Analysis (CCA) versus word label similarity, as shown in Figure \ref{figure:cca-similarity}. It plots the word-level
properties encoded in individual layers. Therefore, higher the number of layers with high similarity better the overall method. Interestingly, the similarity in the later layers increased, which is in line with observations on WavLM. Based on this observation, we selected layers 6 to 11 which exhibit very high CCA similarity uniformly and re-ran the experiments on ASR and PR tasks, which resulted in a performance boost as shown in Table \ref{table:redovc}. These findings suggest that a weighted-sum approach, focusing on layers encoding similar information, is more robust than uniformly using all layers.

\subsection{Comparison with HUBERT using Adapters}
Adapters \citep{chen2023exploringadapters, chen2023chapteradapter} provide a lightweight alternative to fine-tuning pre-trained models to boost the performance on different downstream tasks by adding a minimal number of task-specific parameters. From Table \ref{table:adaptors}, it can be observed that JOOCI performs comparably to HuBERT fine-tuned with adapters, highlighting its capability in learning comprehensive speech representations.
A key limitation of using HuBERT with adapters is that for each new task it requires a new forward pass through the pre-trained model.
While we do not apply adapters to JOOCI in this work, JOOCI does not have this limitation because both the Other and Content embeddings are extracted in parallel during a single forward pass. 

\renewcommand{\arraystretch}{1.2}
\begin{table}[ht]
    \centering
    \caption{Comparison of JOOCI with HuBERT using adapters. FT stands for finetuning. Here 13 is not a mistake and the weighted-sum only uses 13 parameters for each of the 13 layers.}
    \scalebox{0.84}{
    \begin{tabular}{c|c|c|c|c|c}
    \hline
    \textbf{Method} & \textbf{Params} & \textbf{SID} $\uparrow$ & \textbf{ASV} $\downarrow$ & \textbf{ASR} $\downarrow$ & \textbf{PR} $\downarrow$ \\
    \hline
    \hline
    \multicolumn{6}{c}{ \textbf{HuBERT} } \\
    \hline
    FT & 94.68M & 64.56 & 5.15 & 6.35 & 2.45 \\
    \hline
    Weighted-sum & 13 & 81.42 & 5.11 & 6.42 & 5.41\\
    
    \hline
    \hline
    \multicolumn{6}{c}{ \textbf{HuBERT + Adapter} \citep{chen2023chapteradapter}} \\
    \hline
    Houlsby  & 0.60M & 87.71 & 5.29 & 5.88 & 3.00 \\
    \hline
    CHAPTER  & 4.67M & 91.56 & 4.95 & 6.22 & 2.95 \\
    \hline

    \hline
    \hline
    \multicolumn{6}{c}{ \textbf{JOOCI (Ours)} } \\
    \hline
    Weighted-sum & 13 & 90.79 & 4.15 & 5.35 & 4.25 \\
    \hline

    \end{tabular}}
    
    \label{table:adaptors}
\end{table}


\section{Conclusion}
In this paper, we introduced JOOCI, a novel SSL method that successfully learns comprehensive speech representations by jointly optimizing Other (e.g., speaker) and Content (e.g., linguistic) information without compromising the representational-depth for either, unlike WavLM and HuBERT, which enforces layer-wise division. JOOCI addresses this by employing separate encoders for Other and Content, coupled with distinct loss functions, while maintaining a shared encoder for efficient low level feature extraction.
Experimental results on the SUPERB benchmark demonstrate JOOCI’s importance, achieving a 26.5\% improvement over WavLM on jointly maximizing performance on Other (speaker) and Content related tasks and outperforming models of comparable size. Our results validate the importance of using the full representational-depth for building robust representations.

\section{Future Work}
\label{sec:lim}


\textbf{Better Pre-training Methodology :} Employing a different pre-training methodology to learn fine-grained, discriminative features for the Other encoder, such as using neural audio codec latent variables as training targets \citep{defossez2022highencodec}, similar to  \citep{wang2023neuralvalle, chen2024valle2}. 

\textbf{Multi-Talker ASR :} JOOCI is well-suited for multi-talker ASR without requiring any external modules if trained properly. It can be easily adapted by (i) using Other embeddings to classify similar speakers using unsupervised clustering, and (ii) applying the resulting classification as a mask to segment Content embeddings, which can then be used for the ASR task.

\subsubsection*{Acknowledgments}
Hemant Yadav is supported by Microsoft Research India PhD Fellowship program. Rajiv Ratn Shah is partly supported by the Infosys Center for AI, the Center of Design and New Media, and the Center of Excellence in Healthcare at IIIT Delhi.

\newpage

\bibliography{example_paper}

\begin{thebibliography}{48}
\providecommand{\natexlab}[1]{#1}
\providecommand{\url}[1]{\texttt{#1}}
\expandafter\ifx\csname urlstyle\endcsname\relax
  \providecommand{\doi}[1]{doi: #1}\else
  \providecommand{\doi}{doi: \begingroup \urlstyle{rm}\Url}\fi

\bibitem[Alexey(2020)]{alexey2020imagevit}
Alexey, D.
\newblock An image is worth 16x16 words: Transformers for image recognition at scale.
\newblock \emph{arXiv preprint arXiv: 2010.11929}, 2020.

\bibitem[Ao et~al.(2022)Ao, Zhang, Zhou, Liu, Li, Ko, Dai, Li, Qian, and Wei]{ao2022prespeech2c}
Ao, J., Zhang, Z., Zhou, L., Liu, S., Li, H., Ko, T., Dai, L., Li, J., Qian, Y., and Wei, F.
\newblock Pre-training transformer decoder for end-to-end asr model with unpaired speech data.
\newblock \emph{arXiv preprint arXiv:2203.17113}, 2022.

\bibitem[Baevski et~al.(2020)Baevski, Zhou, Mohamed, and Auli]{baevski2020wav2vec2.0}
Baevski, A., Zhou, Y., Mohamed, A., and Auli, M.
\newblock wav2vec 2.0: A framework for self-supervised learning of speech representations.
\newblock \emph{Advances in Neural Information Processing Systems}, 33:\penalty0 12449--12460, 2020.

\bibitem[Baevski et~al.(2022{\natexlab{a}})Baevski, Babu, Hsu, and Auli]{baevski2022efficientdata2vec2.0}
Baevski, A., Babu, A., Hsu, W.-N., and Auli, M.
\newblock Efficient self-supervised learning with contextualized target representations for vision, speech and language.
\newblock \emph{arXiv preprint arXiv:2212.07525}, 2022{\natexlab{a}}.

\bibitem[Baevski et~al.(2022{\natexlab{b}})Baevski, Hsu, Xu, Babu, Gu, and Auli]{baevski2022data2vec}
Baevski, A., Hsu, W.-N., Xu, Q., Babu, A., Gu, J., and Auli, M.
\newblock Data2vec: A general framework for self-supervised learning in speech, vision and language.
\newblock In \emph{International Conference on Machine Learning}, pp.\  1298--1312. PMLR, 2022{\natexlab{b}}.

\bibitem[Borsos et~al.(2023)Borsos, Marinier, Vincent, Kharitonov, Pietquin, Sharifi, Roblek, Teboul, Grangier, Tagliasacchi, et~al.]{borsos2023audiolm}
Borsos, Z., Marinier, R., Vincent, D., Kharitonov, E., Pietquin, O., Sharifi, M., Roblek, D., Teboul, O., Grangier, D., Tagliasacchi, M., et~al.
\newblock Audiolm: a language modeling approach to audio generation.
\newblock \emph{IEEE/ACM transactions on audio, speech, and language processing}, 31:\penalty0 2523--2533, 2023.

\bibitem[Brown et~al.(2020)Brown, Mann, Ryder, Subbiah, Kaplan, Dhariwal, Neelakantan, Shyam, Sastry, Askell, et~al.]{brown2020languagegpt3}
Brown, T., Mann, B., Ryder, N., Subbiah, M., Kaplan, J.~D., Dhariwal, P., Neelakantan, A., Shyam, P., Sastry, G., Askell, A., et~al.
\newblock Language models are few-shot learners.
\newblock \emph{Advances in neural information processing systems}, 33:\penalty0 1877--1901, 2020.

\bibitem[Chan \& Ghosh(2022)Chan and Ghosh]{chan2022contentdisentable2}
Chan, D.~M. and Ghosh, S.
\newblock Content-context factorized representations for automated speech recognition.
\newblock \emph{arXiv preprint arXiv:2205.09872}, 2022.

\bibitem[Chang et~al.(2023{\natexlab{a}})Chang, Liu, and Glass]{chang2023selfspin}
Chang, H.-J., Liu, A.~H., and Glass, J.
\newblock Self-supervised fine-tuning for improved content representations by speaker-invariant clustering.
\newblock \emph{arXiv preprint arXiv:2305.11072}, 2023{\natexlab{a}}.

\bibitem[Chang et~al.(2023{\natexlab{b}})Chang, Liu, and Glass]{chang2023selfspin2}
Chang, H.-J., Liu, A.~H., and Glass, J.
\newblock Self-supervised fine-tuning for improved content representations by speaker-invariant clustering.
\newblock \emph{arXiv preprint arXiv:2305.11072}, 2023{\natexlab{b}}.

\bibitem[Chen et~al.(2022{\natexlab{a}})Chen, Wang, Chen, Wu, Liu, Chen, Li, Kanda, Yoshioka, Xiao, et~al.]{chen2022wavlm}
Chen, S., Wang, C., Chen, Z., Wu, Y., Liu, S., Chen, Z., Li, J., Kanda, N., Yoshioka, T., Xiao, X., et~al.
\newblock Wavlm: Large-scale self-supervised pre-training for full stack speech processing.
\newblock \emph{IEEE Journal of Selected Topics in Signal Processing}, 2022{\natexlab{a}}.

\bibitem[Chen et~al.(2022{\natexlab{b}})Chen, Wu, Wang, Chen, Chen, Liu, Wu, Qian, Wei, Li, et~al.]{chen2022unispeechsat}
Chen, S., Wu, Y., Wang, C., Chen, Z., Chen, Z., Liu, S., Wu, J., Qian, Y., Wei, F., Li, J., et~al.
\newblock Unispeech-sat: Universal speech representation learning with speaker aware pre-training.
\newblock In \emph{ICASSP 2022-2022 IEEE International Conference on Acoustics, Speech and Signal Processing (ICASSP)}, pp.\  6152--6156. IEEE, 2022{\natexlab{b}}.

\bibitem[Chen et~al.(2024)Chen, Liu, Zhou, Liu, Tan, Li, Zhao, Qian, and Wei]{chen2024valle2}
Chen, S., Liu, S., Zhou, L., Liu, Y., Tan, X., Li, J., Zhao, S., Qian, Y., and Wei, F.
\newblock Vall-e 2: Neural codec language models are human parity zero-shot text to speech synthesizers.
\newblock \emph{arXiv preprint arXiv:2406.05370}, 2024.

\bibitem[Chen et~al.(2020)Chen, Kornblith, Norouzi, and Hinton]{chen2020simplesimclr}
Chen, T., Kornblith, S., Norouzi, M., and Hinton, G.
\newblock A simple framework for contrastive learning of visual representations.
\newblock In \emph{International conference on machine learning}, pp.\  1597--1607. PMLR, 2020.

\bibitem[Chen et~al.(2023{\natexlab{a}})Chen, Zheng, Wang, Cheng, and Chen]{chen2023pushing}
Chen, Y., Zheng, S., Wang, H., Cheng, L., and Chen, Q.
\newblock Pushing the limits of self-supervised speaker verification using regularized distillation framework.
\newblock In \emph{ICASSP 2023-2023 IEEE International Conference on Acoustics, Speech and Signal Processing (ICASSP)}, pp.\  1--5. IEEE, 2023{\natexlab{a}}.

\bibitem[Chen et~al.(2023{\natexlab{b}})Chen, Fu, Liu, Li, and Lee]{chen2023exploringadapters}
Chen, Z.-C., Fu, C.-L., Liu, C.-Y., Li, S.-W.~D., and Lee, H.-y.
\newblock Exploring efficient-tuning methods in self-supervised speech models.
\newblock In \emph{2022 IEEE spoken language technology workshop (SLT)}, pp.\  1120--1127. IEEE, 2023{\natexlab{b}}.

\bibitem[Chen et~al.(2023{\natexlab{c}})Chen, Sung, and Lee]{chen2023chapteradapter}
Chen, Z.-C., Sung, Y.-S., and Lee, H.-y.
\newblock Chapter: Exploiting convolutional neural network adapters for self-supervised speech models.
\newblock In \emph{2023 IEEE International Conference on Acoustics, Speech, and Signal Processing Workshops (ICASSPW)}, pp.\  1--5. IEEE, 2023{\natexlab{c}}.

\bibitem[Chung et~al.(2021)Chung, Zhang, Han, Chiu, Qin, Pang, and Wu]{chung2021w2vbert}
Chung, Y.-A., Zhang, Y., Han, W., Chiu, C.-C., Qin, J., Pang, R., and Wu, Y.
\newblock W2v-bert: Combining contrastive learning and masked language modeling for self-supervised speech pre-training.
\newblock In \emph{2021 IEEE Automatic Speech Recognition and Understanding Workshop (ASRU)}, pp.\  244--250. IEEE, 2021.

\bibitem[D{\'e}fossez et~al.(2022)D{\'e}fossez, Copet, Synnaeve, and Adi]{defossez2022highencodec}
D{\'e}fossez, A., Copet, J., Synnaeve, G., and Adi, Y.
\newblock High fidelity neural audio compression.
\newblock \emph{arXiv preprint arXiv:2210.13438}, 2022.

\bibitem[Dubey et~al.(2024)Dubey, Jauhri, Pandey, Kadian, Al-Dahle, Letman, Mathur, Schelten, Yang, Fan, et~al.]{dubey2024llama3}
Dubey, A., Jauhri, A., Pandey, A., Kadian, A., Al-Dahle, A., Letman, A., Mathur, A., Schelten, A., Yang, A., Fan, A., et~al.
\newblock The llama 3 herd of models.
\newblock \emph{arXiv preprint arXiv:2407.21783}, 2024.

\bibitem[Feng et~al.(2022)Feng, Hsu, and Lee]{feng2022silenceissweeter}
Feng, C.-L., Hsu, P.-c., and Lee, H.-y.
\newblock Silence is sweeter than speech: Self-supervised model using silence to store speaker information.
\newblock \emph{arXiv preprint arXiv:2205.03759}, 2022.

\bibitem[Gao et~al.(2019)Gao, Cheng, Zhao, Zhang, Yang, and Torr]{gao2019res2net}
Gao, S.-H., Cheng, M.-M., Zhao, K., Zhang, X.-Y., Yang, M.-H., and Torr, P.
\newblock Res2net: A new multi-scale backbone architecture.
\newblock \emph{IEEE transactions on pattern analysis and machine intelligence}, 43\penalty0 (2):\penalty0 652--662, 2019.

\bibitem[Goodfellow et~al.(2016)Goodfellow, Bengio, and Courville]{Goodfellow-et-al-2016}
Goodfellow, I., Bengio, Y., and Courville, A.
\newblock \emph{Deep Learning}.
\newblock MIT Press, 2016.
\newblock \url{http://www.deeplearningbook.org}.

\bibitem[Grill et~al.(2020)Grill, Strub, Altch{\'e}, Tallec, Richemond, Buchatskaya, Doersch, Avila~Pires, Guo, Gheshlaghi~Azar, et~al.]{grill2020bootstrapbyol}
Grill, J.-B., Strub, F., Altch{\'e}, F., Tallec, C., Richemond, P., Buchatskaya, E., Doersch, C., Avila~Pires, B., Guo, Z., Gheshlaghi~Azar, M., et~al.
\newblock Bootstrap your own latent-a new approach to self-supervised learning.
\newblock \emph{Advances in neural information processing systems}, 33:\penalty0 21271--21284, 2020.

\bibitem[He et~al.(2016)He, Zhang, Ren, and Sun]{resnet}
He, K., Zhang, X., Ren, S., and Sun, J.
\newblock Deep residual learning for image recognition.
\newblock In \emph{Proceedings of the IEEE conference on computer vision and pattern recognition}, pp.\  770--778, 2016.

\bibitem[Hsu et~al.(2021)Hsu, Bolte, Tsai, Lakhotia, Salakhutdinov, and Mohamed]{hsu2021hubert}
Hsu, W.-N., Bolte, B., Tsai, Y.-H.~H., Lakhotia, K., Salakhutdinov, R., and Mohamed, A.
\newblock Hubert: Self-supervised speech representation learning by masked prediction of hidden units.
\newblock \emph{IEEE/ACM Transactions on Audio, Speech, and Language Processing}, 29:\penalty0 3451--3460, 2021.

\bibitem[Mohamed et~al.(2022)Mohamed, Lee, Borgholt, Havtorn, Edin, Igel, Kirchhoff, Li, Livescu, Maal{\o}e, et~al.]{mohamed2022self}
Mohamed, A., Lee, H.-y., Borgholt, L., Havtorn, J.~D., Edin, J., Igel, C., Kirchhoff, K., Li, S.-W., Livescu, K., Maal{\o}e, L., et~al.
\newblock Self-supervised speech representation learning: A review.
\newblock \emph{arXiv preprint arXiv:2205.10643}, 2022.

\bibitem[Okabe et~al.(2018)Okabe, Koshinaka, and Shinoda]{okabe2018attentiveasp}
Okabe, K., Koshinaka, T., and Shinoda, K.
\newblock Attentive statistics pooling for deep speaker embedding.
\newblock \emph{arXiv preprint arXiv:1803.10963}, 2018.

\bibitem[Panayotov et~al.(2015)Panayotov, Chen, Povey, and Khudanpur]{librispeechdataset}
Panayotov, V., Chen, G., Povey, D., and Khudanpur, S.
\newblock Librispeech: An asr corpus based on public domain audio books.
\newblock In \emph{2015 IEEE International Conference on Acoustics, Speech and Signal Processing (ICASSP)}, pp.\  5206--5210, 2015.
\newblock \doi{10.1109/ICASSP.2015.7178964}.

\bibitem[Pasad et~al.(2021)Pasad, Chou, and Livescu]{pasad2021layercpc}
Pasad, A., Chou, J.-C., and Livescu, K.
\newblock Layer-wise analysis of a self-supervised speech representation model.
\newblock In \emph{2021 IEEE Automatic Speech Recognition and Understanding Workshop (ASRU)}, pp.\  914--921. IEEE, 2021.

\bibitem[Pasad et~al.(2023)Pasad, Shi, and Livescu]{pasad2023comparative}
Pasad, A., Shi, B., and Livescu, K.
\newblock Comparative layer-wise analysis of self-supervised speech models.
\newblock In \emph{ICASSP 2023-2023 IEEE International Conference on Acoustics, Speech and Signal Processing (ICASSP)}, pp.\  1--5. IEEE, 2023.

\bibitem[Qian et~al.(2022)Qian, Zhang, Gao, Ni, Lai, Cox, Hasegawa-Johnson, and Chang]{qian2022contentvec}
Qian, K., Zhang, Y., Gao, H., Ni, J., Lai, C.-I., Cox, D., Hasegawa-Johnson, M., and Chang, S.
\newblock Contentvec: An improved self-supervised speech representation by disentangling speakers.
\newblock In \emph{International Conference on Machine Learning}, pp.\  18003--18017. PMLR, 2022.

\bibitem[Riviere et~al.(2020)Riviere, Joulin, Mazar{\'e}, and Dupoux]{modifiedcpc}
Riviere, M., Joulin, A., Mazar{\'e}, P.-E., and Dupoux, E.
\newblock Unsupervised pretraining transfers well across languages.
\newblock In \emph{ICASSP 2020-2020 IEEE International Conference on Acoustics, Speech and Signal Processing (ICASSP)}, pp.\  7414--7418. IEEE, 2020.

\bibitem[Rombach et~al.(2022)Rombach, Blattmann, Lorenz, Esser, and Ommer]{rombach2022highstablediffsion}
Rombach, R., Blattmann, A., Lorenz, D., Esser, P., and Ommer, B.
\newblock High-resolution image synthesis with latent diffusion models.
\newblock In \emph{Proceedings of the IEEE/CVF conference on computer vision and pattern recognition}, pp.\  10684--10695, 2022.

\bibitem[Sang et~al.(2022)Sang, Li, Liu, Arnold, and Wan]{audiodivide}
Sang, M., Li, H., Liu, F., Arnold, A.~O., and Wan, L.
\newblock Self-supervised speaker verification with simple siamese network and self-supervised regularization.
\newblock In \emph{ICASSP 2022-2022 IEEE International Conference on Acoustics, Speech and Signal Processing (ICASSP)}, pp.\  6127--6131. IEEE, 2022.

\bibitem[Schneider et~al.(2019)Schneider, Baevski, Collobert, and Auli]{schneider2019wav2vec}
Schneider, S., Baevski, A., Collobert, R., and Auli, M.
\newblock wav2vec: Unsupervised pre-training for speech recognition.
\newblock \emph{arXiv preprint arXiv:1904.05862}, 2019.

\bibitem[Shi et~al.(2023{\natexlab{a}})Shi, Inaguma, Ma, Kulikov, and Sun]{shi2023MRHuBERT}
Shi, J., Inaguma, H., Ma, X., Kulikov, I., and Sun, A.
\newblock Multi-resolution hubert: Multi-resolution speech self-supervised learning with masked unit prediction.
\newblock \emph{arXiv preprint arXiv:2310.02720}, 2023{\natexlab{a}}.

\bibitem[Shi et~al.(2023{\natexlab{b}})Shi, Tang, Inaguma, GOng, Pino, and Watanabe]{shi2023explorationMRHUBERT}
Shi, J., Tang, Y., Inaguma, H., GOng, H., Pino, J., and Watanabe, S.
\newblock Exploration on hubert with multiple resolutions.
\newblock \emph{arXiv preprint arXiv:2306.01084}, 2023{\natexlab{b}}.

\bibitem[Snyder et~al.(2015)Snyder, Chen, and Povey]{snyder2015musan}
Snyder, D., Chen, G., and Povey, D.
\newblock Musan: A music, speech, and noise corpus.
\newblock \emph{arXiv preprint arXiv:1510.08484}, 2015.

\bibitem[Touvron et~al.(2023)Touvron, Lavril, Izacard, Martinet, Lachaux, Lacroix, Rozi{\`e}re, Goyal, Hambro, Azhar, et~al.]{touvron2023llama1}
Touvron, H., Lavril, T., Izacard, G., Martinet, X., Lachaux, M.-A., Lacroix, T., Rozi{\`e}re, B., Goyal, N., Hambro, E., Azhar, F., et~al.
\newblock Llama: Open and efficient foundation language models.
\newblock \emph{arXiv preprint arXiv:2302.13971}, 2023.

\bibitem[Wang et~al.(2023)Wang, Chen, Wu, Zhang, Zhou, Liu, Chen, Liu, Wang, Li, et~al.]{wang2023neuralvalle}
Wang, C., Chen, S., Wu, Y., Zhang, Z., Zhou, L., Liu, S., Chen, Z., Liu, Y., Wang, H., Li, J., et~al.
\newblock Neural codec language models are zero-shot text to speech synthesizers.
\newblock \emph{arXiv preprint arXiv:2301.02111}, 2023.

\bibitem[Wang et~al.(2018)Wang, Cheng, Liu, and Liu]{wang2018additiveamsoftmax}
Wang, F., Cheng, J., Liu, W., and Liu, H.
\newblock Additive margin softmax for face verification.
\newblock \emph{IEEE Signal Processing Letters}, 25\penalty0 (7):\penalty0 926--930, 2018.

\bibitem[Yadav et~al.(2023)Yadav, Sitaram, and Shah]{yadav2023analysing}
Yadav, H., Sitaram, S., and Shah, R.~R.
\newblock Analysing the masked predictive coding training criterion for pre-training a speech representation model.
\newblock In \emph{ICASSP 2023-2023 IEEE International Conference on Acoustics, Speech and Signal Processing (ICASSP)}, pp.\  1--5. IEEE, 2023.

\bibitem[Yadav et~al.(2024)Yadav, Sitaram, and Shah]{yadav2024ms-hubert}
Yadav, H., Sitaram, S., and Shah, R.~R.
\newblock Ms-hubert: Mitigating pre-training and inference mismatch in masked language modelling methods for learning speech representations.
\newblock \emph{arXiv preprint arXiv:2406.05661}, 2024.

\bibitem[Yang et~al.(2021)Yang, Chi, Chuang, Lai, Lakhotia, Lin, Liu, Shi, Chang, Lin, et~al.]{yang2021superb}
Yang, S.-w., Chi, P.-H., Chuang, Y.-S., Lai, C.-I.~J., Lakhotia, K., Lin, Y.~Y., Liu, A.~T., Shi, J., Chang, X., Lin, G.-T., et~al.
\newblock Superb: Speech processing universal performance benchmark.
\newblock \emph{arXiv preprint arXiv:2105.01051}, 2021.

\bibitem[Yang et~al.(2024)Yang, Chang, Huang, Liu, Lai, Wu, Shi, Chang, Tsai, Huang, et~al.]{yang2024large}
Yang, S.-w., Chang, H.-J., Huang, Z., Liu, A.~T., Lai, C.-I., Wu, H., Shi, J., Chang, X., Tsai, H.-S., Huang, W.-C., et~al.
\newblock A large-scale evaluation of speech foundation models.
\newblock \emph{IEEE/ACM Transactions on Audio, Speech, and Language Processing}, 2024.

\bibitem[Zeghidour et~al.(2021)Zeghidour, Luebs, Omran, Skoglund, and Tagliasacchi]{zeghidour2021soundstream}
Zeghidour, N., Luebs, A., Omran, A., Skoglund, J., and Tagliasacchi, M.
\newblock Soundstream: An end-to-end neural audio codec.
\newblock \emph{IEEE/ACM Transactions on Audio, Speech, and Language Processing}, 30:\penalty0 495--507, 2021.

\bibitem[Zhao et~al.(2023)Zhao, Sun, Lei, Zhu, and Xiong]{zhao2023ccsrddisentangel}
Zhao, X., Sun, H., Lei, Y., Zhu, S., and Xiong, D.
\newblock Ccsrd: Content-centric speech representation disentanglement learning for end-to-end speech translation.
\newblock In \emph{Findings of the Association for Computational Linguistics: EMNLP 2023}, pp.\  5920--5932, 2023.

\end{thebibliography}
\bibliographystyle{icml2025}

\newpage
\appendix
\onecolumn
\section{Appendix}
\subsection{Pre-training}
\label{section:pretraining}

\textbf{Parameter count comparison :} Table \ref{table:param} shows the parameter count of different components of JOOCI during pre-training and inference.



\subsection{SUPERB batch size comparison}
Table \ref{table:batchsizes} shows the comparison of batch sizes used by WavLM and JOOCI. And what is used in the SUPERB benchmark.

\begin{table}[ht]
    \centering
    \begin{tabular}{c|c|c|c}
    \hline
    Task & WavLM & JOOCI & SUPERB\\
    \hline
    SID & 512 & 32 & 32 \\
    \hline
    ASR & 128 & 32 & 32 \\
    \hline
    IC & 128 & 32 & 32 \\
    \hline
    SF & 128 & 32 & 32 \\
    \hline    
    ER & 32 & 32 & 32 \\
    \hline
    \end{tabular}
    \caption{Batchsize used by WavLM and JOOCI and the SUPERB benchmark default.}
\label{table:batchsizes}
\end{table}

\subsection{Weight analysis on the SUPERB benchmark}
\label{section:hubwavsuperb}

\begin{figure}[ht]
\centering
\includegraphics[,width=0.95\columnwidth]{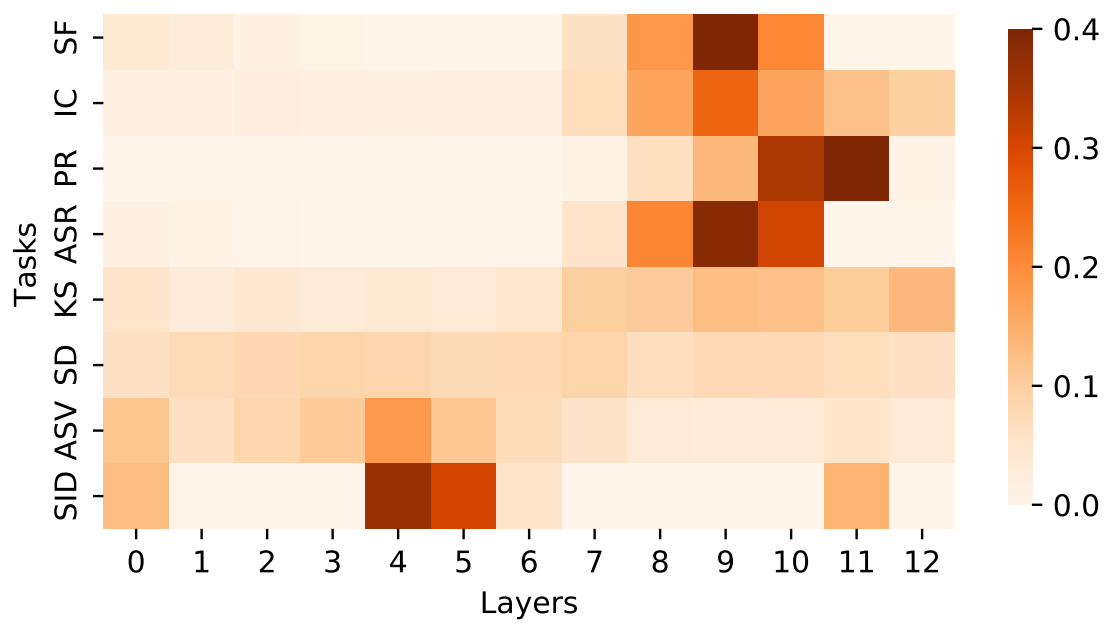}
\caption{HuBERT weight analysis on the SUPERB benchmark. }
\label{figure:hub_superb}
\end{figure}

\begin{figure}[ht]
\centering
\includegraphics[,width=0.95\columnwidth]{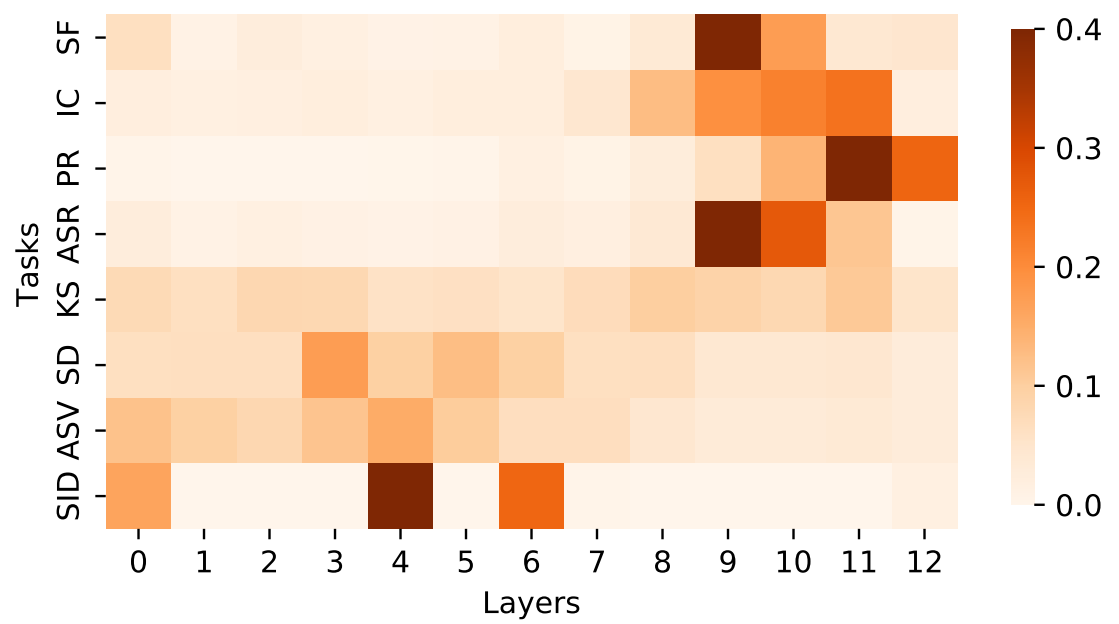}
\caption{WavLM weight analysis on the SUPERB benchmark. }
\label{figure:wav_superb}
\end{figure}

Figure \ref{figure:hub_superb} and Figure \ref{figure:wav_superb} shows the weight analysis for HuBERT and WavLM on the SUPERB benchmark. These images are directly taken from the WavLM paper \citep{chen2022wavlm}.

\end{document}